\documentclass{article}

\usepackage{ijcai19}

\usepackage{times}
\usepackage{soul}
\usepackage{url}
\usepackage{hyperref}
\usepackage[utf8]{inputenc}
\usepackage[small]{caption}
\usepackage{graphicx}
\usepackage{amsmath}
\usepackage{booktabs}
\usepackage{algorithm}
\usepackage{algorithmic}
\usepackage{adjustbox}
\usepackage{graphicx}
\usepackage{url} 
\usepackage{booktabs} 
\usepackage{multirow}
\usepackage{amsmath}
\usepackage{amssymb}
\usepackage{microtype}
\usepackage{textcomp}
\usepackage{etoolbox}
\usepackage{tabu}
\usepackage{array}
\usepackage{tablefootnote}
\usepackage{threeparttable}
\usepackage{pgf-pie}
\usepackage{pgfplots}
\usepackage{array,tabularx}
\usepackage{lipsum}
\usepackage{capt-of}
\usepackage{subfig}

\makeatletter
\patchcmd\@combinedblfloats{\box\@outputbox}{\unvbox\@outputbox}{}{%
  \errmessage{\noexpand\@combinedblfloats could not be patched}%
}%
 \makeatother

\newcommand{\RN}[1]{%
	\textup{\uppercase\expandafter{\romannumeral#1}}%
}
\newcommand*{\MyToprule}{%
  \cmidrule[\heavyrulewidth]%
}
\newcommand*{\MyMidrule}{%
  \cmidrule[\heavyrulewidth]%
}
\newcommand*{\MyBottomrule}{%
  \cmidrule[\heavyrulewidth]%
}

\newcommand\sinquote[1]{
  \textquoteleft #1\textquoteright
}
\newcommand\dblquote[1]{
  \textquotedblleft #1\textquotedblright
}

\newcolumntype{L}[1]{>{\raggedright\let\newline\\\arraybackslash\hspace{0pt}}m{#1}}
\newcolumntype{C}[1]{>{\centering\let\newline\\\arraybackslash\hspace{0pt}}m{#1}}
\newcolumntype{R}[1]{>{\raggedleft\let\newline\\\arraybackslash\hspace{0pt}}m{#1}}

\title{KitcheNette: Predicting and Recommending Food Ingredient Pairings \\ using Siamese Neural Networks}

\author{Donghyeon Park, Keonwoo Kim, Yonggyu Park, Jungwoon Shin\And Jaewoo Kang
\affiliations
Korea University\\
\emails
\{parkdh, akim, yongqyu, jungwoonshin, kangj\}@korea.ac.kr
}

\pgfplotsset{compat=1.14}
\begin{document}

\maketitle

\begin{abstract}
As a vast number of ingredients exist in the culinary world, there are countless food ingredient pairings, but only a small number of pairings have been adopted by chefs and studied by food researchers. In this work, we propose KitcheNette which is a model that predicts food ingredient pairing scores and recommends optimal ingredient pairings. KitcheNette employs Siamese neural networks and is trained on our annotated dataset containing 300K scores of pairings generated from numerous ingredients in food recipes. As the results demonstrate, our model not only outperforms other baseline models but also can recommend complementary food pairings and discover novel ingredient pairings.
\end{abstract}

\section{Introduction}
Many chefs, gourmets, and food-related researchers have focused on studying food pairing for decades. There are books \cite{page2008flavor,dornenburg2009drink} featuring a number of food pairings recommended based on accumulated experiences of professional chefs and food gourmets in the culinary world. Since food pairings are made based on the experiences of experts, food pairing itself is subjective and difficult to quantify. In the academic field, some food-related researchers \cite{ahn2011flavor,ahn2013flavor,garg2017flavordb,simas2017food} focused on determining the qualities of complementary food pairings based on analysis of sharing flavor compounds. However, FlavorDB built by \cite{garg2017flavordb} contains only a limited number of flavor compounds and natural ingredients and a considerable amount of time and effort is required to analyze the flavor compounds of food ingredient.

\begin{figure}[t]
  \centering
  \includegraphics[width=3.375in]{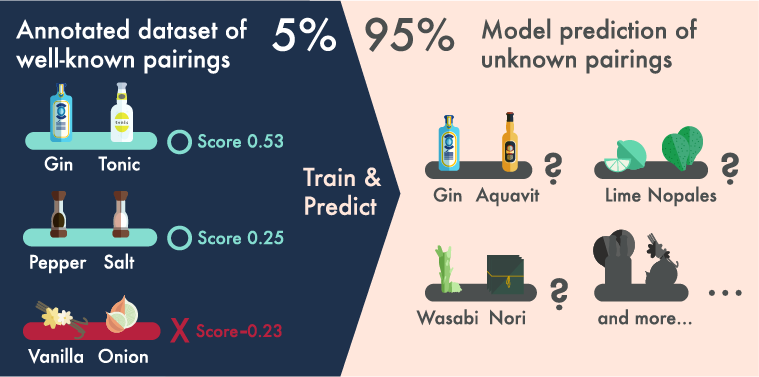}
  \caption{By training on our annotated dataset containing scores of \textit{well-known} food pairings (e.g., gin\&tonic water, salt\&pepper, vanilla\&onion), our model predicts the scores of \textit{unknown} food pairings (e.g., gin\&aquavit, wasabi\&nori, lime\&nopales) that are not annotated because they are less popular or infrequently used.}
  \label{figure:overview}
\end{figure}

In this work, we introduce KitchenNette which is a model based on Siamese neural networks~\cite{koch2015siamese}. As shown in Figure \ref{figure:overview}, KitchenNette first trains on our annotated dataset containing more than 300k scores of \textit{known} pairings, which constitute only 5\% of the total possible number of pairings in our dataset. These quantified scores indicate whether each food pair is complementary or not. Then our trained model predicts the scores of \textit{unknown} pairings consisting of food ingredients that have infrequently or never been used in recipes, which constitute the remaining 95\% of the total number of pairings. The three \textit{unknown} pairings in Figure \ref{figure:overview} that our model found are known to be culture specific pairings in Nordic, Japanese, and Mexican cuisine, respectively. To train our model, we constructed our own dataset which contains the golden standard scores of 300k food ingredient pairings obtained from 1M human-generated cooking recipes~\cite{salvador2017learning,marin2018learning}. Here, the amounts\ and personal preferences of ingredients and the preparation process in cooking were not considered. Our model employs Siamese neural networks with \textit{wide\&deep} architecture designed to learn the relationship of a food pairing. We then conducted experiments to compare it with several baseline models and confirmed that our model KitcheNette outperformed all the other models.

To further evaluate our model's prediction performance, three qualitative analyses were conducted. First, we analyzed some example cases of food pairings to test whether our model successfully predicts the scores of \textit{unknown} pairings. Second, we compared the ranking results of commonly used food ingredient pairings recommended by KitchenNette with those in FlavorDB~\cite{garg2017flavordb}. Our ranking results are more reliable and consistent with human food-pairing knowledge, compared with the results of FlavorDB. Third, we compared the food pairing recommendations of our model, which were based on predicted scores, with those of cooking experts~\cite{page2008flavor,dornenburg2009drink}. We found that most of the recommendations of our model were the same as those of the cooking experts, which demonstrates the accuracy of our model. In addition, our model recommended food pairings with ingredients not commonly featured in recipes. To this extent, our work attempts to broaden the underlying concept of food pairing and introduce a data-driven, deep learning based approach for discovering novel ingredient pairs.
  
The major contributions of our work can be summarized as follows.
\begin{itemize}
\item We propose a data-driven task that discovers new complementary food pairings that have the potential of being used for future recipes.
\item We create a large scale dataset that contains the golden standard scores of food ingredient pairings.
\item We show that KitcheNette\footnote{\textit{\href{https://github.com/dhyeon/KitcheNette}{https://github.com/dmis-lab/KitcheNette}}} which uses Siamese neural networks and \textit{wide\&deep} learning achieves high performance in predicting food ingredient pairing scores.
\item We verify KitchenNette's effectiveness in recommending complementary food pairings\footnote{A demo version of user-interactive KicheNette is available at \textit{\href{http://kitchenette.korea.ac.kr}{http://kitchenette.korea.ac.kr}}.} and discovering new food pairings through qualitative analyses.
\end{itemize}
\section{Related Work}
\subsection{Food Related Research}
\textbf{Researches on Discovering Food Pairings}
\cite{ahn2011flavor} and \cite{ahn2013flavor} introduce a flavor network where the network's edge is built based on the number of flavor compounds shared by culinary ingredients. The flavor network is comprised of 381 ingredients and 1,021 flavor compounds. FlavorDB \cite{garg2017flavordb} combines exisiting food repositories to provide a larger database with the user-interactive page. Food-bridging \cite{simas2017food} improves the flavor network \cite{ahn2011flavor} by adding additional bridges between two ingredients through a chain of pairwise affinities despite the chemical compound similarity of the two ingredients being low. However, they cover only a limited number of flavor compounds and natural ingredients and some well-known food pairings (e.g., red wine and beef) have very few flavor compounds in common. Our work employs a data-driven method to define the scores of food pairings from the human experiences and find the new food pairings in large-scale.

\textbf{Researches on Recommending Recipes}
The Recipe Recommendation~\cite{teng2012recipe} has been proposed to determine whether a food ingredient is essential in a recipe. Recipe Recommendation uses two different recipe networks that can accurately predict recipe ratings. Also, finding a surprising and plausible ingredient combination is not new. \cite{grace2016combining,grace2016surprise} combines cased-based reasoning and deep learning to generate new recipe designs. Our work is similar in that it recommends a new combination of food ingredients, but KitcheNette proposes and trains on a silver standard pairing scores on a larger-scale and is able to suggest novel pairings that never have been tried before.

\subsection{Siamese Neural Networks}

Siamese neural networks~\cite{koch2015siamese} have been employed in various tasks to learn similarities between two different inputs. Also, some variations of Siamese neural networks have been introduced. 

\cite{mueller2016siamese} proposed the Manhattan LSTM model which takes two sentences as input. The Manhattan LSTM model generates vector representations for each sentence and calculates the similarity between the vector representations using the simple function $exp(-||h1 - h2||)$ where $h1$ and $h2$ are the embedding vectors. \cite{yuan2018matching} proposed a customized contrastive loss function that can be divided into a partial loss function for positive pairs and a partial loss function for negative pairs. The loss function widens the distance between two vector representations of negative pairs while narrowing the distance between two vector representations of positive pairs. While these two works (\cite{mueller2016siamese,yuan2018matching}) use Siamese neural networks and take two different inputs of the same type, \cite{liang2018automated}'s proposed Siamese-based model takes one input as the standard by which the other input is evaluated. The input is evaluated based on its similarity to the input used as the standard. Our model is designed to train semantic relationships of a food pair beyond simple distance-based similarity functions.
\section{Dataset}
\subsection{Dataset Description and Preprocessing}
In this work, we utilized Recipe1M~\cite{marin2018learning}, a dataset containing approximately one million recipes and their corresponding images which were collected from multiple popular websites related to cooking. All content in Recipe1M can be divided into two categories: texts and images. The recipe texts of Recipe1M~\cite{marin2018learning} consist of the following two parts: the list of ingredients and the instructions of a recipe. The Im2Recipe~\cite{salvador2017learning} used a bi-directional LSTM based ingredient name extraction module that performs logistic regression on each word in all the lists of ingredients in Recipe1M to extract the ingredient names only apart. For instance,\dblquote{2 tbsp of olive oil} is extracted as olive\_oil. From the instructions for a recipe, \cite{marin2018learning} trains all the vocabulary including ingredient names with the word2vec~\cite{mikolov2013distributed} fashion. Among a total of 30,167 learned vocabulary, we obtained 3,567 unique ingredient names as shown in Table~\ref{table:dataset-statistics}.

\begin{table}[!t]
\centering
\small
\scalebox{0.9}{
    \begin{tabular}{c|r|r|r}
    \MyToprule{1-4}
    \multirow{2}{*}{\textbf{Recipe1M}} & \textbf{\begin{tabular}[c]{@{}c@{}}\# of\\Recipes\end{tabular}} & \textbf{\begin{tabular}[c]{@{}c@{}}\# of \\ Vocab\end{tabular}} & \textbf{\begin{tabular}[c]{@{}c@{}}\# of \\ Ingredient Vocab \end{tabular}} \\\MyMidrule{2-4}
     & 1,029,720 & 30,167 & 3,567 \\\MyMidrule{1-4}
    \multirow{2}{*}{\textbf{\begin{tabular}[c]{@{}c@{}}Ingredient\\ Pairing\\ Dataset\end{tabular}}} & \textbf{\begin{tabular}[c]{@{}c@{}}Total \# of \\ Possible Pairs\end{tabular}} & \textbf{\begin{tabular}[c]{@{}c@{}}\# of \\ \textit{Known} Pairs$^\RN{1}$\end{tabular}} & \textbf{\begin{tabular}[c]{@{}c@{}}\# of \\ \textit{Unknown} Pairs\end{tabular}} \\\MyMidrule{2-4}
    & 6,359,961 & \begin{tabular}[c]{@{}c@{}}356,451\end{tabular} & 6,003,510 \\\MyBottomrule{1-4}
    \end{tabular}
}
\caption{Statistics of Ingredients and Pairings. \textit{Known Pairs}$^\RN{1}$ consist of ingredients whose occurrence counts are greater than 20. In addition, each known pair has a co-occurrence count of at least 5.}
\label{table:dataset-statistics}
\end{table}

\subsection{Food Ingredient Pairing Score Generation}
\subsubsection*{Food Ingredient Pairing Score Dataset}
As mentioned in the earlier sections, traditional ingredient pairing methods have relied on human expertise, such as a long experience in the culinary industry or chemical details of food. As a result, the amount of data to define food pairing and perform deep-learning is absolutely small. To deal with this problem, we propose a new silver standard dataset of food pairing scores that 1) enables training deep-learning models on a large-scale and 2) defines if a pair of food is complementary or not on a scale between -1 and 1. We assumed that the co-occurrence information of ingredients from a large recipe corpus may provide insight into how ingredients are combined in each recipe. Within the scope of this study, we did not consider amounts of ingredients nor their cooking procedures since our dataset is based on a statistical co-occurrence information.

\subsubsection*{Normalized Point-wise Mutual Information}
We calculated our golden standard food pairing scores based on point-wise mutual information (PMI) as introduced in \cite{teng2012recipe}. The PMI score \eqref{eq:pmi} is the probability of two elements co-occurring $p(x,y)$, which is compared to the probability of each element occurring separately $p(x)$, $p(y)$. The custom score is designed to accurately represent good/bad pairs by penalizing highly popular ingredients such as salt or butter with low co-occurrence pairs. On the other hand, a pair that shows high co-occurrence with less popular ingredients will represent a good, meaningful pair (e.g., wasabi\&nori).

\scriptsize
\begin{equation}\label{eq:pmi}
\operatorname{pmi}(x;y) = \log\frac{p(x,y)}{p(x)p(y)}
\end{equation}
\textnormal{where:}
\begin{equation*}
p(x,y) = \frac{\textnormal{\# of recipes where x and y occur together}}{\textnormal{\# of recipes}} \\
\end{equation*}
\begin{equation*}
p(x) = \frac{\textnormal{\# of recipes where x occurs}}{\textnormal{\# of recipes}} \\
\end{equation*}
\begin{equation*}
p(y) = \frac{\textnormal{\# of recipes where y occurs}}{\textnormal{\# of recipes}} \\
\end{equation*}
\normalsize

In our work, we used the normalized version of PMI (NPMI~\cite{bouma2009normalized}) to better train and fit our regression model. Point-wise mutual information can be normalized between -1 and +1 where -1 (in the limit) is given for never occurring together, 0 for independence, and +1 for complete co-occurrence. Thus, the scores between -1 and 1 intuitively determine if the pair is well suited or not.

\scriptsize
\begin{equation}\label{eq:npmi}
\operatorname{npmi}(x;y) = \frac{\operatorname{pmi}(x;y)}{h(x, y)}
\end{equation}
where:
\begin{equation*}
h(x, y) = -\log{p(x,y)}\\
\end{equation*}
\normalsize

\begin{table}[!t]
    \centering
    \scalebox{0.75}{
    \begin{tabular}{r|r|c|c}
    \toprule[1.5pt]
    \textbf{\begin{tabular}[c]{@{}c@{}}Ingredient1 {[}Count{]}\end{tabular}} & \textbf{\begin{tabular}[c]{@{}c@{}}Ingredient2 {[}Count{]}\end{tabular}} & \textbf{\begin{tabular}[c]{@{}c@{}}Co-\\ occurrence\end{tabular}} & \textbf{\begin{tabular}[c]{@{}c@{}}Pairing\\score\end{tabular}}\\\midrule
    \multirow{10}{*}{\begin{tabular}[c]{@{}c@{}}vanilla {[}51,756{]}\end{tabular}} & \begin{tabular}[c]{@{}c@{}}baking\_soda {[}58,931{]}\end{tabular} & 14,657 & 0.376 \\
     & \begin{tabular}[c]{@{}c@{}}cocoa {[}6,520{]}\end{tabular} & 2,759 & 0.360 \\
     & \begin{tabular}[c]{@{}c@{}}powdered\_sugar {[}26,729{]}\end{tabular} & 6,558 & 0.314 \\
     & \begin{tabular}[c]{@{}c@{}}nut {[}9,090{]}\end{tabular} & 2,865 & 0.312 \\
     & \begin{tabular}[c]{@{}c@{}}chocolate\_chips {[}9,172{]}\end{tabular} & 2,821 & 0.307 \\\MyMidrule{2-4}
     & \begin{tabular}[c]{@{}c@{}}onion {[}191,691{]}\end{tabular} & 12 & -0.589 \\
     & \begin{tabular}[c]{@{}c@{}}soy\_sauce {[}40,518{]}\end{tabular} & 6 & -0.483 \\
     & \begin{tabular}[c]{@{}c@{}}salt\_and\_pepper {[}46,534{]}\end{tabular} & 14 & -0.479 \\
     & \begin{tabular}[c]{@{}c@{}}garlic {[}46,534{]}\end{tabular} & 9 & -0.477 \\
     & \begin{tabular}[c]{@{}c@{}}pepper {[}68,984{]}\end{tabular} & 26 & -0.462 \\\bottomrule[1.5pt]
    \end{tabular}
    }
    \caption{Ingredient Pairing Dataset. The five best\&worst ingredients in our dataset for pairing with vanilla. The \textit{The Flavor Bible}~\protect\cite{page2008flavor} recommended pairing chocolate, coffee, cream, ice cream, or sugar with vanilla.}\label{table:dataset-example}
\end{table}

\subsubsection*{Generating Ingredient Pairing Dataset based on NPMI Scores}
Ideally, we would calculate all the food ingredient pairing scores of 6,359,961 possible pairing${3,567\choose 2}$ generated from 3,567 unique ingredients. However, we found that ingredients that rarely occurred in 1 million recipe texts may act as noisy samples. Also, ingredients that rarely co-occur may lower the performance of the model. Therefore, we removed ingredients whose occurrence count does not exceed 20 and ingredient pairings whose co-occurrence count dose not exceed 5 to build our golden standard \textit{known} pairs in Table~\ref{table:dataset-statistics}.

As a result, we obtained a total of 356,451 valid \textit{known} pairings. All the other pairings were considered as \textit{unknown} pairings. The final distribution of food ingredient pairing scores follows the approximated normal distribution. We assume that the ingredient pairing scores in the upper 5\% ($\mu+2\sigma$) of the distribution are the top scores and scores lower than 0.274 are in the lower distribution. The scores of the five best and worst ingredients to pair with vanilla are contained in our dataset, as shown in Table \ref{table:dataset-example}.
\section{Model}

\subsection{Learning Ingredient Representations}
We propose a model that predicts the scores of ingredient pairings. Our model architecture consists of two major components, as shown in Figure \ref{figure:kitchenette}. The first is the \sinquote{Ingredient Representation Component} which uses Siamese neural networks~\cite{koch2015siamese} where two identical multi-layer perceptrons (MLPs) with the same weights each receive a different 300-dimensional word vector representation. Each MLP has two fully connected layers which process the input ingredient vector. Let $(X_a, X_b)$ be a pair of ingredients represented as 300-dimensional word vectors. $W_*$ and $b_*$ are the shared weights and bias of an MLP, respectively, and $f(\cdot)$ denotes the activation function for non-linearity. We use $f(\cdot)$ ($\text{max}(x, 0)$) as rectified linear units (ReLUs). The learned representations $(h_a, h_b)$ of this pair are mathematically expressed as follows:

\begin{figure}[!t]
  \centering
  \includegraphics[width=3.375in, height=3.125in]{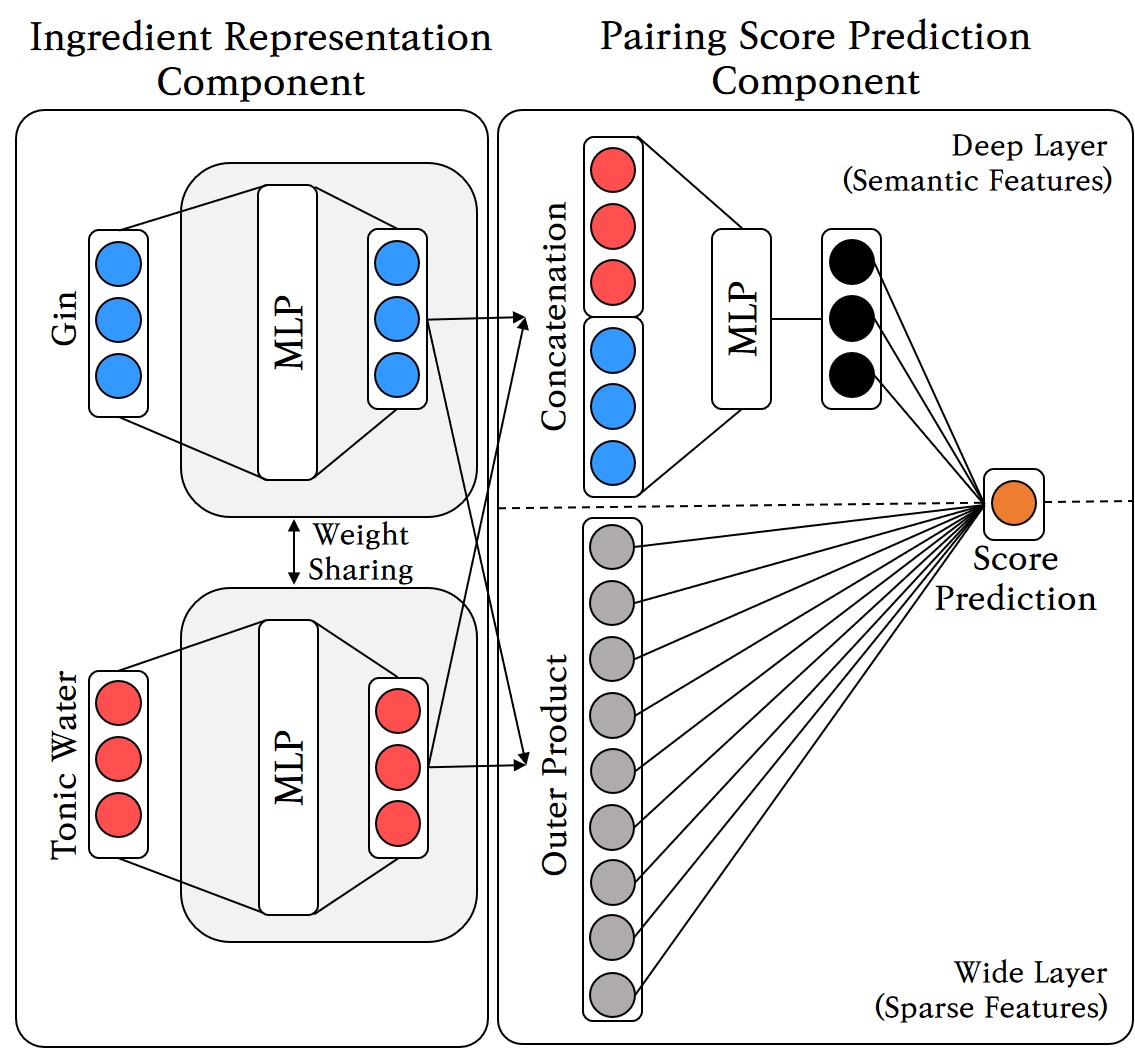}
  \caption{Overview of our KitcheNette architecture}
  \label{figure:kitchenette}
\end{figure}

\scriptsize
\begin{equation*}
h_a = f(W_2f(W_1 X_a + b_1) + b_2)
\end{equation*}
\begin{equation*}
h_b = f(W_2f(W_1 X_b + b_1) + b_2)
\end{equation*}
\normalsize

where $W_1 \in \mathbb{R}^{i \times 300}$, $W_2 \in \mathbb{R}^{j \times i}$, $b_1 \in \mathbb{R}^{i}$, and $b_2 \in \mathbb{R}^{j}$, and $i$ and $j$ are the number of hidden units.

\subsection{Predicting Food Ingredient Pairing Scores}
In the \sinquote{Pairing Score Prediction Component}, we employ \textit{wide\&deep} learning~\cite{cheng2016wide}. The layer is divided into a \textit{wide layer} and a \textit{deep layer}. In the \textit{deep layer}, two $j$-dimensional learned representation vectors are concatenated and passed to another MLP that computes a joint representation of two ingredients. This joint representation is denoted as deep vector $d$ and is mathematically expressed as follows:

\scriptsize
\begin{equation*}
d = f(W_4f(W_3 (h_a,h_b) + b_3) + b_4)
\end{equation*}
\normalsize

where $W_3 \in \mathbb{R}^{j \times 2j}$, $W_4 \in \mathbb{R}^{j \times j}$, $b_3 \in \mathbb{R}^{j}$, $b_4 \in \mathbb{R}^{j}$, and $j$ is the number of hidden units in each layer. In the \textit{wide layer}, the outer product of two $j$-dimensional learned representation vectors is computed and a ${j}\times{j}$ matrix is generated. The matrix is then flattened to a ${j^2}$-dimensional \textit{wide vector} $w$ which is mathematically expressed as follows:

\scriptsize
\begin{equation*}
w =g(h_a \otimes h_b)
\end{equation*}
\normalsize

where g denotes a flattening function that converts a ${n}\times{n}$ matrix into a ${n^2}$-dimensional single vector. 

The \textit{wide vector} $w$ is then directly concatenated to the \textit{deep vector} $d$. The concatenation of the wide and deep vectors is passed to the last fully connected layer to compute the pair score for the final output. Overall, as $(w, d)$ is the concatenation of the \textit{wide} and \textit{deep} vectors, the final output score $Y$ for the ingredient pair $(X_a, X_b)$ is mathematically calculated as follows:

\scriptsize
\begin{equation*}
Y = W_5 (w,d) + b_5
\end{equation*}
\normalsize

where $W_5\in \mathbb{R}^{1 \times (j^2+j)}$ and $b_5 \in \mathbb{R}^{1}$, and $j$ is the number of hidden units in each layer.

\subsection{Model Training Details}

We train our proposed model to minimize the loss function (Mean Squared Error) which can be expressed as follows:  

\scriptsize
\begin{equation*}
L(\Theta) = \frac{1}{N}\sum_{a, b}(y_{ab} - Y_{ab})^2
\end{equation*}
\normalsize

where L is the computed loss function to be minimized during training, $\Theta$ are the model parameters to be trained, $y_{ab}$ is the true score value, $Y_{ab}$ is the predicted score value, and N is the total number of input pairs used for training. We use the Adam optimizer for our model.

\section{Experiment}
\begin{table*}[!t]
    \centering
    \scalebox{0.85}{
    \begin{tabular}{c|rrrrr|rrrrr}
    \toprule[1pt]
     \multirow{2}{*}{Model} & \multicolumn{5}{c|}{Validation} & \multicolumn{5}{c}{Test} \\
     & RMSE & MSE & MAE & CORR & R2 & RMSE & MSE & MAE & CORR & R2 \\\midrule[1pt]
     Cosine Similarity & - & - & - & - & - & 0.1802 & 0.0325 & 0.1328 & 0.3952 & -1.6026 
     \\\midrule[0.1pt]
     Gradient Boosting & 0.1073 & 0.0115 & 0.0815 & 0.3339 & 0.0773 & 0.1073 & 0.0115 & 0.0815 & 0.3351 & 0.0776 \\
     SGD & 0.0993 & 0.0099 & 0.0762 & 0.4585 & 0.2102 & 0.0984 & 0.0097 & 0.0759 & 0.4730 & 0.2236 \\
     Linear SVR & 0.0993 & 0.0099 & 0.0762 & 0.4588 & 0.2105 & 0.0984 & 0.0097 & 0.0759 & 0.4731 & 0.2238 \\
     Random Forest & 0.0802 & 0.0064 & 0.0612 & 0.7015 & 0.4846 & 0.0799 & 0.0064 & 0.0611 & 0.7042 & 0.4885 \\
     Extra Tree & 0.0742 & 0.0055 & 0.0566 & 0.7664 & 0.5586 & 0.0738 & 0.0054 & 0.0563 & 0.7689 & 0.5637
     \\\midrule[0.1pt]
     Siamese Network & 0.0726 & 0.0054 & 0.0540 & 0.8223 & 0.5679 & 0.0729 & 0.0054 & 0.0544 & 0.8235 & 0.5662 \\
     \textbf{KitcheNette} & \textbf{0.0421} & \textbf{0.0018} & \textbf{0.0320} & \textbf{0.9249} & \textbf{0.8551} & \textbf{0.0417} & \textbf{0.0018} & \textbf{0.0317} & \textbf{0.9266} & \textbf{0.8583} \\\bottomrule[1pt]
    \end{tabular}
    }
    \caption{Prediction results of the models.}\label{table:regression}
\end{table*}

\subsection{Baseline Models}
We first evaluated the baseline models before evaluating our proposed model. We first predicted the pairing scores by simply calculating the cosine similarity between two input ingredient vectors. We employed the following machine learning models from the Python Scikit-learn~\cite{scikit-learn} package as our baseline models: Linear Support Vector Regressor, Random Forest Regressor, Extra Tree Regressor, SGD Regressor, and Gradient Boosting. Additionally, a simple version of Siamese Neural Network~\cite{koch2015siamese} is used one of the baseline models. All these models are fitted with hyperparameters estimated by the built-in grid search.

\subsection{Main Results}
As illustrated in Table \ref{table:regression}, the following five metrics were utilized to evaluate model performance: Root Mean Square Error (RMSE), Mean Square Error (MSE), Mean Absolute Error (MAE), Correlation (CORR), and R squared (R2). Our KitcheNette model clearly outperforms the baseline models in all metrics. 

\begin{figure}
\centering
\subfloat[NDCG@K]{\label{figure:ndcg}
\resizebox{4.25cm}{!}{
\begin{tikzpicture}
\begin{axis}[
    xlabel={NDCG},
    ylabel={Accuracy},
    xmin=-0.5, xmax=5.5,
    ymin=-0.05, ymax=1,
    xtick={0.0, 1.0, 2.0, 3.0, 4.0, 5.0},
    xticklabels={10,20,50,100,500,1000},
    ytick={0,0.20,0.40,0.60,0.80,1.00},
    legend columns=-1,
    legend style={at={(0.5,-0.2)},anchor=north},
    legend cell align=left,
    ymajorgrids=true,
    grid style=dashed,
]
 
\addplot[
    color=blue,
    mark=*,
    ]
    coordinates {
    (0,0.328)(1,0.39)(2,0.546)(3,0.536)(4,0.675)(5,0.694 )
    };
\addplot[
    color=red,
    mark=square,
    ]
    coordinates {
    (0,0.155)(1,0.204)(2,0.188)(3,0.192)(4,0.387)(5,0.447)
    };
\addplot[
    color=orange,
    mark=otimes,
    ]
    coordinates {
    (0,0)(1,0.079)(2,0.128)(3,0.188)(4,0.330)(5,0.395)
    };
\addplot[
    color=purple,
    mark=triangle,
    ]
    coordinates {
    (0,0)(1,0)(2,0.08)(3,0.088)(4,0.233)(5,0.290)
    };
\addplot[
    color=green,
    mark=diamond,
    ]
    coordinates {
    (0,0)(1,0)(2,0.08)(3,0.088)(4,0.233)(5,0.291)
    };
\addplot[
    color=yellow,
    mark=otimes,
    ]
    coordinates {
    (0,0)(1,0)(2,0)(3,0)(4,0.083)(5,0.120)
    };
\legend{KitcheNette,ET,RF,SVR,SGD,GB}
\end{axis}
\end{tikzpicture}
}
}%
\subfloat[ROC Curve]{\label{figure:roc}
\resizebox{4.25cm}{!}{
\begin{tikzpicture}
\begin{axis}[
    xlabel={False Positive Rate (\%)},
    ylabel={True Positive Rate (\%)},
    xmin=-0.05, xmax=1.05,
    ymin=-0.05, ymax=1.05,
    xtick={0,0.20,0.40,0.60,0.80,1.00},
    ytick={0,0.20,0.40,0.60,0.80,1.00},
    legend columns=-1,
    legend style={at={(0.5,-0.2)},anchor=north},
    legend cell align=left,
    ymajorgrids=true,
    grid style=dashed,
]
\addplot[
    color=blue,
    ]
    coordinates {
    (0.0,0.0)(0.0,0.139)(0.0,0.188)(0.0,0.21)(0.0,0.252)(0.0,0.262)(0.0,0.282)(0.001,0.29)(0.001,0.31)(0.001,0.326)(0.001,0.345)(0.001,0.361)(0.001,0.372)(0.001,0.382)(0.001,0.391)(0.001,0.401)(0.001,0.412)(0.002,0.42)(0.002,0.432)(0.002,0.439)(0.002,0.451)(0.002,0.459)(0.002,0.464)(0.002,0.468)(0.002,0.478)(0.002,0.485)(0.003,0.491)(0.003,0.499)(0.003,0.507)(0.003,0.519)(0.003,0.527)(0.003,0.533)(0.003,0.537)(0.004,0.541)(0.004,0.546)(0.004,0.55)(0.004,0.556)(0.004,0.562)(0.004,0.566)(0.004,0.571)(0.004,0.578)(0.004,0.583)(0.005,0.588)(0.005,0.59)(0.005,0.593)(0.005,0.599)(0.005,0.601)(0.005,0.605)(0.005,0.611)(0.005,0.615)(0.006,0.621)(0.006,0.624)(0.006,0.629)(0.006,0.632)(0.006,0.637)(0.007,0.641)(0.007,0.648)(0.007,0.651)(0.007,0.654)(0.007,0.657)(0.007,0.663)(0.008,0.665)(0.008,0.668)(0.008,0.672)(0.008,0.676)(0.009,0.679)(0.009,0.683)(0.009,0.685)(0.009,0.688)(0.009,0.693)(0.009,0.696)(0.009,0.699)(0.01,0.701)(0.01,0.707)(0.01,0.71)(0.01,0.714)(0.01,0.718)(0.011,0.721)(0.011,0.726)(0.011,0.731)(0.011,0.734)(0.011,0.737)(0.011,0.741)(0.012,0.743)(0.012,0.745)(0.012,0.749)(0.012,0.752)(0.013,0.754)(0.013,0.757)(0.013,0.76)(0.013,0.763)(0.014,0.766)(0.014,0.768)(0.014,0.771)(0.014,0.774)(0.014,0.776)(0.014,0.778)(0.015,0.781)(0.015,0.784)(0.015,0.786)(0.016,0.788)(0.016,0.791)(0.016,0.795)(0.016,0.798)(0.017,0.8)(0.017,0.802)(0.018,0.805)(0.018,0.807)(0.018,0.81)(0.018,0.812)(0.018,0.815)(0.019,0.817)(0.019,0.82)(0.019,0.822)(0.02,0.825)(0.02,0.827)(0.02,0.83)(0.021,0.832)(0.021,0.835)(0.022,0.837)(0.022,0.84)(0.022,0.842)(0.022,0.845)(0.023,0.847)(0.023,0.85)(0.023,0.852)(0.024,0.855)(0.024,0.858)(0.025,0.861)(0.025,0.863)(0.026,0.865)(0.026,0.869)(0.027,0.871)(0.027,0.873)(0.028,0.876)(0.028,0.879)(0.029,0.881)(0.03,0.883)(0.03,0.886)(0.03,0.888)(0.031,0.89)(0.031,0.893)(0.032,0.895)(0.032,0.898)(0.033,0.9)(0.033,0.902)(0.034,0.905)(0.035,0.907)(0.036,0.91)(0.036,0.912)(0.037,0.914)(0.038,0.916)(0.04,0.919)(0.04,0.921)(0.041,0.923)(0.042,0.925)(0.043,0.927)(0.044,0.929)(0.045,0.932)(0.046,0.934)(0.047,0.937)(0.048,0.939)(0.051,0.941)(0.052,0.943)(0.053,0.945)(0.055,0.947)(0.056,0.949)(0.058,0.951)(0.06,0.953)(0.061,0.955)(0.063,0.957)(0.065,0.959)(0.068,0.961)(0.069,0.963)(0.071,0.966)(0.075,0.968)(0.077,0.97)(0.083,0.972)(0.087,0.973)(0.098,0.975)(0.1,0.977)(0.104,0.979)(0.109,0.982)(0.113,0.983)(0.125,0.985)(0.132,0.987)(0.145,0.989)(0.156,0.991)(0.172,0.991)(0.18,0.993)(0.209,0.994)(0.245,0.995)(0.271,0.997)(0.308,0.998)(0.325,0.998)(0.38,1.0)(0.462,1.0)(0.475,1.0)(0.646,1.0)(0.8,1.0)(0.861,1.0)(0.928,1.0)
    };
\addplot[
    color=red,
    ]
    coordinates {
    (0.0,0.001)(0.0,0.042)(0.0,0.05)(0.0,0.072)(0.0,0.084)(0.0,0.092)(0.0,0.106)(0.001,0.117)(0.001,0.127)(0.001,0.138)(0.001,0.145)(0.001,0.152)(0.001,0.163)(0.001,0.171)(0.001,0.179)(0.002,0.184)(0.002,0.196)(0.002,0.202)(0.002,0.207)(0.002,0.219)(0.002,0.227)(0.002,0.232)(0.002,0.235)(0.003,0.247)(0.003,0.252)(0.003,0.258)(0.003,0.264)(0.003,0.268)(0.003,0.275)(0.004,0.278)(0.004,0.283)(0.004,0.289)(0.004,0.295)(0.004,0.3)(0.004,0.304)(0.004,0.311)(0.005,0.317)(0.005,0.323)(0.005,0.327)(0.005,0.338)(0.005,0.341)(0.006,0.347)(0.006,0.35)(0.006,0.355)(0.006,0.361)(0.006,0.364)(0.007,0.369)(0.007,0.373)(0.007,0.377)(0.007,0.38)(0.008,0.384)(0.008,0.389)(0.008,0.392)(0.008,0.396)(0.008,0.4)(0.009,0.404)(0.009,0.408)(0.009,0.413)(0.009,0.419)(0.009,0.425)(0.01,0.429)(0.01,0.433)(0.01,0.437)(0.01,0.44)(0.011,0.444)(0.011,0.449)(0.011,0.452)(0.011,0.457)(0.012,0.46)(0.012,0.463)(0.013,0.468)(0.013,0.471)(0.013,0.475)(0.013,0.479)(0.014,0.486)(0.014,0.491)(0.014,0.493)(0.014,0.497)(0.015,0.5)(0.015,0.504)(0.015,0.508)(0.016,0.513)(0.016,0.516)(0.016,0.519)(0.016,0.522)(0.017,0.527)(0.017,0.53)(0.017,0.536)(0.018,0.539)(0.018,0.542)(0.018,0.546)(0.019,0.549)(0.019,0.553)(0.019,0.556)(0.019,0.559)(0.02,0.563)(0.02,0.568)(0.02,0.571)(0.021,0.576)(0.021,0.58)(0.021,0.584)(0.022,0.587)(0.022,0.591)(0.023,0.594)(0.023,0.599)(0.023,0.602)(0.023,0.605)(0.024,0.608)(0.024,0.612)(0.025,0.615)(0.025,0.618)(0.026,0.622)(0.026,0.625)(0.026,0.628)(0.027,0.631)(0.027,0.634)(0.028,0.637)(0.029,0.641)(0.029,0.645)(0.029,0.649)(0.03,0.652)(0.03,0.655)(0.031,0.658)(0.031,0.661)(0.032,0.664)(0.033,0.667)(0.033,0.67)(0.034,0.674)(0.034,0.678)(0.035,0.682)(0.035,0.685)(0.036,0.69)(0.036,0.693)(0.037,0.696)(0.037,0.698)(0.037,0.701)(0.038,0.704)(0.038,0.707)(0.039,0.71)(0.039,0.714)(0.041,0.718)(0.042,0.721)(0.043,0.723)(0.043,0.726)(0.044,0.73)(0.045,0.732)(0.045,0.736)(0.046,0.739)(0.047,0.742)(0.048,0.745)(0.049,0.748)(0.05,0.751)(0.051,0.755)(0.052,0.758)(0.053,0.761)(0.054,0.764)(0.054,0.767)(0.056,0.769)(0.058,0.774)(0.059,0.777)(0.059,0.78)(0.06,0.783)(0.061,0.785)(0.063,0.788)(0.064,0.791)(0.065,0.794)(0.066,0.797)(0.067,0.8)(0.068,0.803)(0.069,0.805)(0.07,0.808)(0.072,0.811)(0.074,0.814)(0.074,0.817)(0.075,0.82)(0.077,0.822)(0.079,0.826)(0.08,0.829)(0.082,0.832)(0.084,0.834)(0.085,0.837)(0.087,0.84)(0.088,0.843)(0.089,0.846)(0.092,0.849)(0.095,0.852)(0.096,0.855)(0.099,0.858)(0.1,0.861)(0.102,0.864)(0.105,0.867)(0.106,0.871)(0.108,0.874)(0.112,0.877)(0.114,0.88)(0.119,0.883)(0.121,0.886)(0.123,0.888)(0.126,0.891)(0.13,0.894)(0.133,0.897)(0.138,0.9)(0.14,0.903)(0.146,0.906)(0.152,0.908)(0.155,0.912)(0.158,0.915)(0.161,0.918)(0.166,0.921)(0.176,0.924)(0.178,0.927)(0.182,0.929)(0.187,0.932)(0.194,0.935)(0.207,0.938)(0.216,0.941)(0.229,0.944)(0.235,0.946)(0.247,0.949)(0.254,0.952)(0.258,0.955)(0.27,0.958)(0.29,0.96)(0.297,0.962)(0.316,0.965)(0.33,0.968)(0.346,0.971)(0.362,0.974)(0.387,0.977)(0.415,0.979)(0.449,0.98)(0.486,0.983)(0.539,0.985)(0.594,0.987)(0.617,0.99)(0.635,0.991)(0.673,0.994)(0.726,0.997)(0.822,0.999)
    };
\addplot[
    color=orange,
    ]
    coordinates {
    (0.0,0.0)(0.0,0.027)(0.0,0.039)(0.0,0.054)(0.0,0.064)(0.0,0.082)(0.001,0.09)(0.001,0.098)(0.001,0.103)(0.001,0.11)(0.001,0.115)(0.001,0.12)(0.001,0.125)(0.001,0.129)(0.001,0.134)(0.002,0.142)(0.002,0.148)(0.002,0.152)(0.002,0.154)(0.002,0.161)(0.002,0.167)(0.002,0.172)(0.002,0.175)(0.003,0.18)(0.003,0.182)(0.003,0.185)(0.003,0.189)(0.003,0.194)(0.003,0.2)(0.004,0.204)(0.004,0.207)(0.004,0.21)(0.004,0.213)(0.004,0.217)(0.004,0.223)(0.005,0.226)(0.005,0.229)(0.005,0.233)(0.005,0.235)(0.005,0.238)(0.005,0.242)(0.005,0.249)(0.006,0.254)(0.006,0.256)(0.006,0.26)(0.006,0.267)(0.006,0.269)(0.006,0.271)(0.006,0.275)(0.007,0.279)(0.007,0.281)(0.007,0.285)(0.007,0.289)(0.008,0.291)(0.008,0.294)(0.008,0.297)(0.008,0.3)(0.008,0.304)(0.008,0.307)(0.009,0.31)(0.009,0.313)(0.009,0.316)(0.009,0.318)(0.009,0.323)(0.01,0.326)(0.01,0.33)(0.01,0.334)(0.01,0.336)(0.01,0.339)(0.01,0.344)(0.011,0.346)(0.011,0.35)(0.011,0.352)(0.011,0.355)(0.012,0.359)(0.012,0.361)(0.012,0.366)(0.013,0.372)(0.013,0.374)(0.013,0.376)(0.013,0.379)(0.013,0.382)(0.014,0.386)(0.014,0.388)(0.014,0.391)(0.014,0.393)(0.015,0.396)(0.015,0.401)(0.015,0.403)(0.015,0.405)(0.016,0.408)(0.016,0.41)(0.016,0.413)(0.016,0.416)(0.016,0.419)(0.017,0.421)(0.017,0.424)(0.017,0.427)(0.018,0.43)(0.018,0.432)(0.018,0.435)(0.019,0.437)(0.019,0.44)(0.019,0.444)(0.02,0.446)(0.02,0.451)(0.02,0.453)(0.02,0.455)(0.021,0.457)(0.021,0.46)(0.021,0.463)(0.022,0.465)(0.022,0.468)(0.022,0.471)(0.023,0.474)(0.023,0.477)(0.023,0.48)(0.023,0.482)(0.024,0.485)(0.024,0.487)(0.024,0.489)(0.024,0.492)(0.025,0.495)(0.025,0.498)(0.026,0.501)(0.026,0.504)(0.026,0.509)(0.026,0.511)(0.027,0.514)(0.027,0.517)(0.028,0.519)(0.028,0.522)(0.028,0.524)(0.028,0.527)(0.029,0.53)(0.029,0.532)(0.029,0.535)(0.029,0.539)(0.03,0.541)(0.03,0.544)(0.03,0.546)(0.031,0.549)(0.031,0.552)(0.032,0.554)(0.032,0.557)(0.032,0.559)(0.033,0.562)(0.033,0.564)(0.033,0.566)(0.034,0.569)(0.035,0.571)(0.035,0.574)(0.036,0.576)(0.036,0.578)(0.036,0.58)(0.037,0.583)(0.037,0.585)(0.037,0.588)(0.038,0.59)(0.038,0.592)(0.039,0.595)(0.039,0.597)(0.039,0.6)(0.04,0.602)(0.04,0.604)(0.041,0.606)(0.041,0.608)(0.042,0.61)(0.042,0.613)(0.042,0.615)(0.042,0.618)(0.042,0.62)(0.043,0.623)(0.043,0.625)(0.044,0.627)(0.044,0.629)(0.045,0.632)(0.046,0.634)(0.046,0.637)(0.046,0.639)(0.047,0.641)(0.047,0.644)(0.047,0.646)(0.048,0.648)(0.049,0.651)(0.049,0.653)(0.05,0.655)(0.05,0.658)(0.051,0.66)(0.052,0.662)(0.052,0.664)(0.053,0.666)(0.053,0.668)(0.054,0.671)(0.055,0.673)(0.055,0.675)(0.056,0.677)(0.058,0.679)(0.058,0.681)(0.059,0.683)(0.06,0.686)(0.06,0.689)(0.061,0.691)(0.062,0.693)(0.062,0.696)(0.063,0.698)(0.064,0.7)(0.065,0.702)(0.065,0.705)(0.066,0.707)(0.067,0.709)(0.068,0.711)(0.07,0.713)(0.071,0.715)(0.071,0.717)(0.072,0.72)(0.073,0.722)(0.074,0.724)(0.075,0.726)(0.076,0.728)(0.078,0.73)(0.078,0.732)(0.079,0.734)(0.08,0.736)(0.081,0.739)(0.082,0.741)(0.084,0.743)(0.085,0.745)(0.087,0.747)(0.088,0.749)(0.089,0.752)(0.09,0.754)(0.091,0.756)(0.092,0.758)(0.093,0.76)(0.094,0.762)(0.096,0.764)(0.098,0.767)(0.099,0.769)(0.099,0.771)(0.1,0.773)(0.102,0.775)(0.103,0.778)(0.104,0.78)(0.105,0.783)(0.106,0.785)(0.108,0.787)(0.109,0.789)(0.111,0.791)(0.112,0.793)(0.113,0.795)(0.115,0.798)(0.115,0.8)(0.118,0.802)(0.119,0.804)(0.121,0.806)(0.122,0.808)(0.124,0.81)(0.125,0.812)(0.13,0.815)(0.131,0.817)(0.132,0.819)(0.134,0.821)(0.136,0.824)(0.137,0.826)(0.139,0.828)(0.14,0.83)(0.142,0.832)(0.143,0.834)(0.144,0.837)(0.146,0.839)(0.149,0.841)(0.151,0.843)(0.153,0.845)(0.156,0.847)(0.159,0.849)(0.16,0.851)(0.164,0.854)(0.167,0.856)(0.17,0.858)(0.173,0.86)(0.175,0.862)(0.178,0.864)(0.179,0.866)(0.182,0.868)(0.185,0.87)(0.187,0.872)(0.189,0.874)(0.193,0.876)(0.197,0.878)(0.203,0.88)(0.205,0.883)(0.21,0.885)(0.214,0.887)(0.218,0.889)(0.225,0.891)(0.227,0.893)(0.231,0.895)(0.239,0.897)(0.241,0.899)(0.245,0.901)(0.247,0.903)(0.252,0.905)(0.255,0.907)(0.259,0.91)(0.265,0.912)(0.271,0.914)(0.277,0.916)(0.282,0.918)(0.285,0.92)(0.297,0.922)(0.298,0.924)(0.305,0.926)(0.314,0.928)(0.324,0.93)(0.328,0.932)(0.337,0.935)(0.343,0.937)(0.348,0.939)(0.353,0.941)(0.374,0.943)(0.377,0.945)(0.382,0.947)(0.393,0.949)(0.4,0.951)(0.415,0.953)(0.425,0.955)(0.436,0.957)(0.457,0.959)(0.47,0.961)(0.475,0.963)(0.494,0.965)(0.531,0.967)(0.544,0.969)(0.556,0.971)(0.566,0.973)(0.58,0.975)(0.605,0.976)(0.627,0.978)(0.637,0.98)(0.651,0.982)(0.664,0.984)(0.688,0.985)(0.72,0.987)(0.751,0.988)(0.763,0.99)(0.779,0.991)(0.8,0.992)(0.814,0.994)(0.841,0.996)(0.909,0.998)(0.944,0.999)(0.986,1.0)
    };
\addplot[
    color=purple,
    ]
    coordinates {
    (0.0,0.0)(0.0,0.009)(0.0,0.014)(0.0,0.022)(0.0,0.028)(0.0,0.032)(0.001,0.038)(0.001,0.043)(0.001,0.049)(0.001,0.054)(0.001,0.058)(0.001,0.064)(0.001,0.069)(0.002,0.073)(0.002,0.076)(0.002,0.082)(0.002,0.087)(0.002,0.092)(0.002,0.095)(0.002,0.101)(0.003,0.105)(0.003,0.109)(0.003,0.112)(0.003,0.115)(0.003,0.12)(0.003,0.123)(0.004,0.128)(0.004,0.13)(0.004,0.133)(0.004,0.137)(0.004,0.14)(0.004,0.143)(0.005,0.146)(0.005,0.148)(0.005,0.151)(0.005,0.154)(0.006,0.159)(0.006,0.163)(0.006,0.166)(0.006,0.168)(0.006,0.171)(0.007,0.175)(0.007,0.184)(0.007,0.186)(0.007,0.189)(0.007,0.192)(0.008,0.194)(0.008,0.196)(0.008,0.2)(0.008,0.203)(0.008,0.206)(0.008,0.21)(0.009,0.213)(0.009,0.216)(0.009,0.219)(0.009,0.221)(0.009,0.225)(0.009,0.227)(0.01,0.23)(0.01,0.235)(0.01,0.239)(0.01,0.242)(0.011,0.244)(0.011,0.247)(0.011,0.251)(0.011,0.254)(0.012,0.256)(0.012,0.258)(0.012,0.261)(0.013,0.263)(0.013,0.266)(0.013,0.268)(0.014,0.272)(0.014,0.275)(0.014,0.278)(0.015,0.281)(0.015,0.283)(0.015,0.286)(0.015,0.289)(0.016,0.291)(0.016,0.295)(0.016,0.298)(0.017,0.3)(0.018,0.304)(0.018,0.306)(0.018,0.309)(0.019,0.311)(0.019,0.314)(0.02,0.316)(0.021,0.318)(0.021,0.32)(0.021,0.322)(0.021,0.324)(0.021,0.327)(0.022,0.33)(0.022,0.333)(0.023,0.335)(0.023,0.339)(0.023,0.341)(0.024,0.343)(0.025,0.345)(0.025,0.347)(0.025,0.349)(0.025,0.352)(0.026,0.354)(0.026,0.356)(0.026,0.359)(0.027,0.363)(0.027,0.365)(0.028,0.369)(0.028,0.371)(0.028,0.373)(0.028,0.376)(0.029,0.378)(0.029,0.381)(0.03,0.384)(0.03,0.386)(0.031,0.388)(0.031,0.39)(0.032,0.393)(0.032,0.395)(0.033,0.398)(0.033,0.4)(0.034,0.403)(0.034,0.405)(0.034,0.408)(0.035,0.41)(0.035,0.412)(0.036,0.415)(0.036,0.417)(0.037,0.419)(0.037,0.421)(0.037,0.423)(0.038,0.425)(0.038,0.427)(0.039,0.43)(0.039,0.432)(0.04,0.435)(0.04,0.438)(0.041,0.44)(0.042,0.442)(0.042,0.445)(0.043,0.447)(0.043,0.449)(0.044,0.451)(0.044,0.453)(0.045,0.456)(0.046,0.459)(0.046,0.462)(0.047,0.464)(0.047,0.467)(0.048,0.47)(0.049,0.473)(0.05,0.475)(0.051,0.477)(0.052,0.48)(0.052,0.482)(0.053,0.484)(0.053,0.486)(0.054,0.488)(0.055,0.492)(0.056,0.494)(0.056,0.496)(0.057,0.499)(0.057,0.501)(0.058,0.503)(0.058,0.505)(0.059,0.507)(0.06,0.51)(0.06,0.512)(0.061,0.515)(0.062,0.517)(0.064,0.519)(0.064,0.521)(0.065,0.523)(0.066,0.525)(0.066,0.528)(0.067,0.53)(0.068,0.532)(0.069,0.534)(0.07,0.537)(0.072,0.539)(0.073,0.541)(0.074,0.544)(0.075,0.546)(0.076,0.548)(0.076,0.55)(0.077,0.552)(0.078,0.555)(0.079,0.557)(0.081,0.559)(0.082,0.562)(0.083,0.564)(0.083,0.566)(0.084,0.568)(0.085,0.571)(0.086,0.573)(0.087,0.575)(0.088,0.578)(0.089,0.58)(0.09,0.582)(0.092,0.584)(0.093,0.586)(0.093,0.588)(0.094,0.59)(0.095,0.593)(0.096,0.595)(0.096,0.598)(0.097,0.6)(0.098,0.602)(0.1,0.604)(0.101,0.606)(0.101,0.608)(0.102,0.61)(0.103,0.612)(0.105,0.615)(0.106,0.617)(0.108,0.619)(0.109,0.621)(0.112,0.624)(0.113,0.627)(0.115,0.629)(0.116,0.631)(0.117,0.633)(0.118,0.635)(0.119,0.637)(0.121,0.639)(0.122,0.641)(0.123,0.643)(0.124,0.645)(0.128,0.647)(0.129,0.649)(0.13,0.651)(0.131,0.654)(0.133,0.656)(0.134,0.658)(0.136,0.661)(0.137,0.663)(0.138,0.665)(0.14,0.667)(0.141,0.669)(0.142,0.672)(0.144,0.674)(0.145,0.676)(0.146,0.678)(0.148,0.68)(0.148,0.682)(0.15,0.684)(0.153,0.686)(0.154,0.689)(0.156,0.691)(0.158,0.693)(0.16,0.695)(0.161,0.697)(0.162,0.699)(0.164,0.701)(0.165,0.703)(0.166,0.705)(0.168,0.708)(0.17,0.71)(0.173,0.712)(0.176,0.715)(0.179,0.717)(0.179,0.719)(0.181,0.721)(0.182,0.723)(0.187,0.724)(0.19,0.727)(0.191,0.729)(0.192,0.731)(0.194,0.733)(0.196,0.735)(0.199,0.737)(0.2,0.739)(0.202,0.741)(0.204,0.744)(0.205,0.746)(0.207,0.748)(0.211,0.75)(0.213,0.752)(0.217,0.754)(0.219,0.756)(0.221,0.758)(0.224,0.76)(0.226,0.762)(0.227,0.764)(0.228,0.766)(0.231,0.768)(0.231,0.771)(0.234,0.773)(0.235,0.775)(0.238,0.777)(0.243,0.779)(0.246,0.781)(0.248,0.784)(0.25,0.786)(0.253,0.789)(0.257,0.791)(0.264,0.793)(0.268,0.795)(0.273,0.798)(0.276,0.8)(0.28,0.802)(0.283,0.804)(0.288,0.806)(0.292,0.808)(0.296,0.81)(0.299,0.812)(0.303,0.815)(0.306,0.817)(0.309,0.819)(0.313,0.821)(0.314,0.823)(0.317,0.825)(0.319,0.827)(0.323,0.829)(0.328,0.831)(0.333,0.833)(0.334,0.835)(0.339,0.837)(0.342,0.839)(0.347,0.841)(0.351,0.844)(0.354,0.846)(0.356,0.848)(0.361,0.85)(0.366,0.852)(0.369,0.854)(0.373,0.856)(0.375,0.858)(0.38,0.86)(0.384,0.862)(0.387,0.864)(0.393,0.866)(0.399,0.868)(0.401,0.87)(0.405,0.872)(0.41,0.874)(0.415,0.876)(0.422,0.878)(0.43,0.88)(0.434,0.883)(0.439,0.885)(0.446,0.887)(0.45,0.889)(0.455,0.891)(0.46,0.893)(0.464,0.895)(0.469,0.897)(0.479,0.899)(0.484,0.901)(0.5,0.903)(0.509,0.905)(0.513,0.907)(0.528,0.909)(0.536,0.911)(0.541,0.913)(0.55,0.915)(0.556,0.917)(0.558,0.919)(0.566,0.921)(0.569,0.923)(0.576,0.925)(0.584,0.927)(0.591,0.929)(0.603,0.932)(0.61,0.934)(0.617,0.936)(0.632,0.938)(0.646,0.94)(0.65,0.941)(0.657,0.944)(0.663,0.946)(0.667,0.948)(0.676,0.95)(0.691,0.952)(0.704,0.954)(0.72,0.956)(0.722,0.958)(0.733,0.96)(0.742,0.962)(0.753,0.964)(0.769,0.966)(0.776,0.968)(0.782,0.97)(0.799,0.972)(0.809,0.974)(0.825,0.976)(0.837,0.977)(0.851,0.979)(0.869,0.981)(0.887,0.983)(0.912,0.985)(0.933,0.988)(0.947,0.989)(0.96,0.991)(0.974,0.993)(0.98,0.995)(0.993,0.998)(0.999,1.0)
    };
\addplot[
    color=green,
    ]
    coordinates {
    (0.0,0.0)(0.0,0.009)(0.0,0.017)(0.0,0.024)(0.0,0.032)(0.001,0.038)(0.001,0.041)(0.001,0.045)(0.001,0.048)(0.001,0.052)(0.001,0.059)(0.001,0.063)(0.001,0.071)(0.002,0.074)(0.002,0.077)(0.002,0.082)(0.002,0.087)(0.002,0.092)(0.002,0.098)(0.003,0.102)(0.003,0.105)(0.003,0.109)(0.003,0.112)(0.003,0.116)(0.003,0.119)(0.004,0.122)(0.004,0.125)(0.004,0.129)(0.004,0.132)(0.004,0.137)(0.004,0.139)(0.005,0.141)(0.005,0.148)(0.005,0.154)(0.005,0.157)(0.005,0.16)(0.006,0.163)(0.006,0.167)(0.006,0.171)(0.006,0.175)(0.006,0.178)(0.007,0.182)(0.007,0.186)(0.007,0.189)(0.007,0.193)(0.007,0.195)(0.008,0.198)(0.008,0.204)(0.008,0.206)(0.008,0.208)(0.009,0.211)(0.009,0.215)(0.009,0.218)(0.009,0.222)(0.009,0.227)(0.01,0.229)(0.01,0.233)(0.01,0.236)(0.01,0.238)(0.011,0.241)(0.011,0.244)(0.011,0.246)(0.011,0.249)(0.011,0.251)(0.012,0.254)(0.012,0.256)(0.012,0.259)(0.012,0.261)(0.013,0.264)(0.013,0.268)(0.014,0.27)(0.014,0.273)(0.014,0.278)(0.015,0.28)(0.015,0.283)(0.015,0.285)(0.015,0.287)(0.016,0.289)(0.016,0.292)(0.016,0.295)(0.017,0.298)(0.017,0.3)(0.018,0.303)(0.018,0.305)(0.018,0.308)(0.019,0.31)(0.019,0.313)(0.02,0.315)(0.02,0.318)(0.02,0.32)(0.021,0.322)(0.021,0.324)(0.022,0.327)(0.022,0.329)(0.023,0.332)(0.023,0.335)(0.024,0.339)(0.024,0.342)(0.025,0.344)(0.025,0.346)(0.025,0.349)(0.025,0.351)(0.026,0.354)(0.026,0.356)(0.026,0.359)(0.027,0.362)(0.027,0.365)(0.027,0.367)(0.028,0.37)(0.028,0.373)(0.028,0.375)(0.029,0.377)(0.029,0.38)(0.03,0.382)(0.03,0.385)(0.03,0.387)(0.031,0.389)(0.031,0.391)(0.031,0.393)(0.032,0.395)(0.032,0.398)(0.033,0.4)(0.034,0.402)(0.034,0.405)(0.035,0.407)(0.035,0.41)(0.036,0.412)(0.036,0.415)(0.036,0.417)(0.037,0.42)(0.037,0.422)(0.038,0.424)(0.038,0.427)(0.039,0.429)(0.039,0.431)(0.039,0.433)(0.04,0.436)(0.041,0.438)(0.042,0.441)(0.042,0.444)(0.043,0.446)(0.043,0.448)(0.044,0.451)(0.044,0.453)(0.045,0.455)(0.046,0.46)(0.046,0.463)(0.047,0.465)(0.048,0.467)(0.049,0.469)(0.049,0.472)(0.05,0.474)(0.051,0.477)(0.051,0.48)(0.052,0.482)(0.052,0.484)(0.053,0.487)(0.054,0.489)(0.055,0.492)(0.055,0.494)(0.055,0.496)(0.056,0.498)(0.056,0.5)(0.057,0.502)(0.057,0.504)(0.058,0.506)(0.059,0.508)(0.059,0.51)(0.06,0.512)(0.061,0.515)(0.062,0.517)(0.062,0.519)(0.063,0.522)(0.064,0.524)(0.065,0.527)(0.066,0.529)(0.068,0.531)(0.069,0.533)(0.07,0.535)(0.072,0.538)(0.073,0.54)(0.074,0.542)(0.075,0.544)(0.075,0.546)(0.077,0.549)(0.078,0.551)(0.079,0.554)(0.079,0.556)(0.08,0.558)(0.081,0.56)(0.082,0.562)(0.083,0.564)(0.083,0.566)(0.084,0.568)(0.084,0.57)(0.085,0.572)(0.087,0.575)(0.088,0.577)(0.089,0.579)(0.09,0.581)(0.091,0.583)(0.092,0.586)(0.092,0.588)(0.094,0.59)(0.095,0.592)(0.096,0.594)(0.096,0.597)(0.097,0.599)(0.098,0.601)(0.099,0.603)(0.1,0.605)(0.101,0.607)(0.103,0.61)(0.104,0.612)(0.105,0.615)(0.107,0.617)(0.108,0.619)(0.109,0.621)(0.11,0.623)(0.111,0.625)(0.113,0.628)(0.114,0.63)(0.115,0.632)(0.116,0.634)(0.117,0.637)(0.119,0.639)(0.121,0.641)(0.123,0.644)(0.125,0.646)(0.126,0.648)(0.127,0.651)(0.128,0.653)(0.129,0.655)(0.132,0.657)(0.134,0.659)(0.135,0.661)(0.136,0.663)(0.137,0.665)(0.139,0.668)(0.14,0.67)(0.141,0.671)(0.144,0.673)(0.146,0.676)(0.146,0.678)(0.148,0.68)(0.149,0.682)(0.152,0.685)(0.153,0.687)(0.155,0.689)(0.155,0.691)(0.158,0.694)(0.159,0.696)(0.161,0.698)(0.164,0.7)(0.165,0.702)(0.167,0.705)(0.17,0.707)(0.172,0.709)(0.174,0.711)(0.175,0.713)(0.178,0.715)(0.179,0.717)(0.18,0.719)(0.182,0.721)(0.183,0.723)(0.185,0.726)(0.188,0.728)(0.19,0.73)(0.193,0.732)(0.196,0.734)(0.197,0.737)(0.2,0.739)(0.201,0.741)(0.204,0.743)(0.206,0.745)(0.207,0.747)(0.209,0.749)(0.211,0.751)(0.214,0.754)(0.215,0.756)(0.219,0.758)(0.222,0.76)(0.223,0.762)(0.226,0.764)(0.229,0.766)(0.232,0.768)(0.234,0.771)(0.235,0.773)(0.24,0.775)(0.242,0.777)(0.243,0.779)(0.246,0.781)(0.25,0.783)(0.252,0.785)(0.255,0.787)(0.258,0.79)(0.262,0.792)(0.264,0.794)(0.266,0.796)(0.273,0.798)(0.279,0.8)(0.285,0.802)(0.288,0.804)(0.29,0.806)(0.293,0.808)(0.295,0.81)(0.298,0.812)(0.3,0.814)(0.304,0.816)(0.305,0.818)(0.308,0.82)(0.314,0.822)(0.315,0.824)(0.319,0.827)(0.322,0.829)(0.325,0.831)(0.331,0.833)(0.332,0.835)(0.336,0.837)(0.34,0.839)(0.345,0.841)(0.354,0.843)(0.357,0.845)(0.361,0.847)(0.365,0.85)(0.368,0.852)(0.372,0.854)(0.375,0.856)(0.38,0.858)(0.384,0.86)(0.392,0.863)(0.395,0.865)(0.397,0.867)(0.403,0.869)(0.405,0.871)(0.409,0.873)(0.413,0.875)(0.42,0.877)(0.425,0.879)(0.435,0.881)(0.439,0.883)(0.448,0.886)(0.452,0.888)(0.455,0.89)(0.459,0.892)(0.462,0.894)(0.473,0.896)(0.481,0.898)(0.486,0.9)(0.49,0.902)(0.503,0.905)(0.509,0.907)(0.519,0.909)(0.53,0.911)(0.54,0.913)(0.551,0.915)(0.56,0.917)(0.563,0.919)(0.572,0.921)(0.576,0.923)(0.579,0.925)(0.595,0.927)(0.603,0.929)(0.609,0.932)(0.616,0.934)(0.619,0.936)(0.635,0.938)(0.642,0.94)(0.647,0.942)(0.657,0.944)(0.662,0.946)(0.668,0.948)(0.682,0.95)(0.691,0.952)(0.711,0.954)(0.715,0.956)(0.728,0.959)(0.736,0.961)(0.742,0.963)(0.756,0.965)(0.766,0.967)(0.78,0.968)(0.787,0.971)(0.8,0.973)(0.812,0.975)(0.828,0.976)(0.84,0.978)(0.851,0.979)(0.866,0.981)(0.88,0.983)(0.909,0.985)(0.93,0.988)(0.945,0.99)(0.964,0.992)(0.974,0.994)(0.984,0.996)(0.994,0.998)
    };
\addplot[
    color=yellow,
    ]
    coordinates {
    (0.0,0.0)(0.004,0.063)(0.007,0.081)(0.01,0.103)(0.013,0.12)(0.015,0.132)(0.018,0.139)(0.02,0.152)(0.024,0.168)(0.025,0.177)(0.029,0.202)(0.031,0.209)(0.032,0.222)(0.035,0.235)(0.037,0.238)(0.039,0.249)(0.044,0.26)(0.046,0.272)(0.05,0.289)(0.054,0.297)(0.057,0.304)(0.058,0.309)(0.06,0.314)(0.062,0.326)(0.064,0.331)(0.067,0.34)(0.071,0.348)(0.074,0.351)(0.081,0.366)(0.085,0.371)(0.088,0.377)(0.091,0.387)(0.093,0.392)(0.094,0.395)(0.098,0.409)(0.1,0.415)(0.102,0.418)(0.113,0.436)(0.126,0.453)(0.129,0.456)(0.131,0.458)(0.134,0.464)(0.135,0.471)(0.138,0.476)(0.141,0.48)(0.144,0.487)(0.146,0.49)(0.151,0.497)(0.156,0.5)(0.16,0.509)(0.166,0.515)(0.181,0.532)(0.191,0.543)(0.205,0.551)(0.208,0.557)(0.209,0.56)(0.214,0.566)(0.216,0.569)(0.222,0.578)(0.225,0.583)(0.234,0.595)(0.237,0.6)(0.237,0.602)(0.249,0.608)(0.261,0.616)(0.263,0.617)(0.263,0.618)(0.265,0.624)(0.267,0.628)(0.268,0.629)(0.271,0.634)(0.277,0.636)(0.28,0.637)(0.285,0.639)(0.293,0.643)(0.297,0.646)(0.303,0.652)(0.311,0.656)(0.312,0.657)(0.313,0.658)(0.315,0.663)(0.318,0.667)(0.319,0.67)(0.321,0.673)(0.323,0.676)(0.329,0.68)(0.331,0.682)(0.339,0.689)(0.341,0.694)(0.343,0.697)(0.343,0.7)(0.344,0.704)(0.348,0.707)(0.35,0.71)(0.355,0.711)(0.356,0.715)(0.358,0.716)(0.361,0.717)(0.362,0.717)(0.363,0.719)(0.365,0.72)(0.366,0.722)(0.367,0.723)(0.368,0.726)(0.369,0.729)(0.372,0.732)(0.373,0.735)(0.376,0.74)(0.377,0.742)(0.38,0.745)(0.382,0.749)(0.385,0.751)(0.389,0.754)(0.39,0.756)(0.391,0.757)(0.392,0.759)(0.393,0.76)(0.395,0.761)(0.396,0.765)(0.398,0.767)(0.399,0.768)(0.401,0.772)(0.404,0.776)(0.405,0.778)(0.408,0.78)(0.412,0.783)(0.414,0.787)(0.418,0.789)(0.434,0.795)(0.435,0.795)(0.441,0.798)(0.447,0.801)(0.448,0.802)(0.449,0.804)(0.449,0.804)(0.45,0.805)(0.451,0.805)(0.452,0.807)(0.453,0.807)(0.454,0.808)(0.456,0.81)(0.459,0.815)(0.461,0.815)(0.465,0.82)(0.466,0.82)(0.467,0.821)(0.474,0.825)(0.474,0.825)(0.477,0.826)(0.482,0.83)(0.483,0.83)(0.488,0.834)(0.49,0.836)(0.493,0.836)(0.496,0.838)(0.501,0.84)(0.503,0.841)(0.503,0.842)(0.505,0.845)(0.506,0.846)(0.507,0.847)(0.508,0.848)(0.509,0.85)(0.512,0.851)(0.516,0.853)(0.516,0.853)(0.518,0.853)(0.521,0.854)(0.522,0.854)(0.526,0.855)(0.533,0.859)(0.534,0.859)(0.545,0.864)(0.548,0.865)(0.549,0.866)(0.55,0.866)(0.552,0.868)(0.555,0.869)(0.558,0.872)(0.56,0.872)(0.561,0.873)(0.565,0.874)(0.566,0.875)(0.569,0.878)(0.57,0.878)(0.571,0.878)(0.571,0.879)(0.573,0.88)(0.575,0.881)(0.577,0.884)(0.581,0.884)(0.582,0.884)(0.582,0.885)(0.59,0.888)(0.592,0.889)(0.594,0.891)(0.597,0.892)(0.598,0.892)(0.599,0.892)(0.602,0.893)(0.603,0.895)(0.605,0.896)(0.615,0.898)(0.617,0.898)(0.618,0.899)(0.623,0.9)(0.624,0.901)(0.626,0.901)(0.629,0.901)(0.63,0.902)(0.631,0.902)(0.632,0.902)(0.633,0.904)(0.634,0.904)(0.635,0.904)(0.636,0.905)(0.637,0.905)(0.637,0.905)(0.638,0.906)(0.64,0.906)(0.641,0.906)(0.642,0.907)(0.648,0.911)(0.65,0.912)(0.651,0.912)(0.652,0.912)(0.653,0.912)(0.654,0.913)(0.658,0.913)(0.666,0.92)(0.668,0.92)(0.669,0.92)(0.669,0.92)(0.68,0.927)(0.681,0.927)(0.686,0.929)(0.688,0.931)(0.689,0.932)(0.697,0.936)(0.699,0.937)(0.699,0.937)(0.7,0.938)(0.702,0.939)(0.703,0.939)(0.705,0.94)(0.705,0.941)(0.707,0.942)(0.707,0.942)(0.708,0.942)(0.709,0.943)(0.71,0.943)(0.712,0.944)(0.712,0.944)(0.713,0.944)(0.714,0.944)(0.715,0.944)(0.721,0.947)(0.722,0.947)(0.722,0.947)(0.723,0.947)(0.724,0.947)(0.725,0.947)(0.73,0.949)(0.731,0.949)(0.744,0.954)(0.746,0.955)(0.747,0.956)(0.748,0.956)(0.757,0.959)(0.76,0.96)(0.761,0.96)(0.768,0.963)(0.768,0.963)(0.769,0.964)(0.77,0.965)(0.77,0.965)(0.771,0.965)(0.774,0.966)(0.777,0.966)(0.78,0.966)(0.79,0.967)(0.791,0.968)(0.793,0.968)(0.794,0.968)(0.794,0.968)(0.795,0.968)(0.797,0.969)(0.798,0.969)(0.8,0.971)(0.801,0.971)(0.802,0.971)(0.803,0.971)(0.806,0.972)(0.807,0.972)(0.812,0.973)(0.813,0.973)(0.814,0.974)(0.815,0.974)(0.816,0.974)(0.816,0.974)(0.818,0.974)(0.82,0.975)(0.821,0.975)(0.822,0.975)(0.824,0.976)(0.824,0.976)(0.827,0.977)(0.83,0.978)(0.832,0.978)(0.835,0.978)(0.836,0.979)(0.837,0.979)(0.838,0.979)(0.838,0.979)(0.843,0.98)(0.854,0.98)(0.854,0.981)(0.855,0.981)(0.856,0.982)(0.856,0.982)(0.858,0.982)(0.859,0.982)(0.86,0.982)(0.861,0.982)(0.863,0.983)(0.867,0.983)(0.868,0.983)(0.868,0.983)(0.883,0.987)(0.883,0.987)(0.884,0.987)(0.885,0.988)(0.886,0.988)(0.887,0.988)(0.887,0.988)(0.889,0.988)(0.889,0.988)(0.89,0.988)(0.902,0.989)(0.903,0.989)(0.905,0.989)(0.907,0.989)(0.908,0.989)(0.909,0.989)(0.909,0.989)(0.912,0.989)(0.914,0.99)(0.915,0.991)(0.915,0.991)(0.916,0.991)(0.916,0.991)(0.925,0.991)(0.925,0.992)(0.926,0.992)(0.927,0.993)(0.928,0.993)(0.929,0.993)(0.93,0.993)(0.932,0.994)(0.932,0.994)(0.932,0.994)(0.934,0.994)(0.934,0.994)(0.935,0.994)(0.939,0.994)(0.939,0.994)(0.94,0.994)(0.94,0.994)(0.942,0.995)(0.946,0.995)(0.947,0.995)(0.948,0.995)(0.948,0.995)(0.948,0.996)(0.949,0.997)(0.95,0.997)(0.952,0.997)(0.956,0.998)(0.96,0.998)(0.961,0.998)(0.962,0.998)(0.962,0.998)(0.963,0.998)(0.972,0.998)(0.973,0.998)(0.974,0.998)(0.974,0.998)(0.979,0.998)(0.98,0.999)(0.981,0.999)(0.985,0.999)(0.986,0.999)(0.986,0.999)(0.998,1.0)(1.0,1.0)
    };
\addplot[
    color=black,
    dashed,
    ]
    coordinates {
    (0.0,0.0)(0.2,0.2)(0.4,0.4)(0.6,0.6)(0.8,0.8)(1.0,1.0)
    };    
\legend{KitcheNette, ET, RF, SVR, SGD, GB}
\end{axis}
\end{tikzpicture}
}
}%
\hfill 
\caption{Additional Model Prediction Results.}\label{figure:results}
\end{figure}
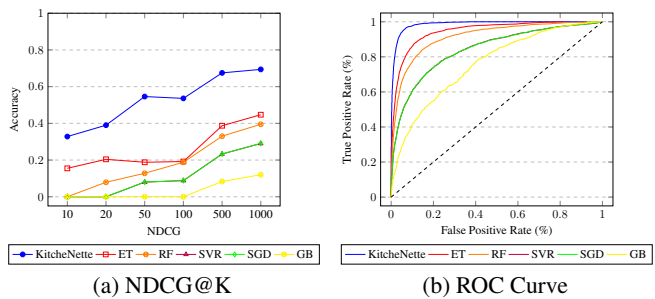

We use Normalized Discounted Cumulative Gain (NDCG@K) to evaluate the ranking performance of our model (Figure \ref{figure:ndcg}) and employ ROC curve to evaluate the sensitivity of our model (Figure \ref{figure:roc}). In terms of NDCG@K, our model outperforms all the baseline models in making accurate predictions. The ROC curve is also used to measure the classification performance of the models. As mentioned in Section 3.2, we regarded all pairings with prediction scores of 0.274 or higher as complementary pairings; pairings with lower scores were considered non-complementary. The ROC curve results demonstrate that our KitcheNette model achieves higher performance than all the other models in predicting complementary pairings. 

\subsection{Ablation Study}
We performed ablation tests to evaluate each feature of KitcheNette. As illustrated in Table \ref{table:ablation}, the \textit{wide\&deep} architecture and ingredient embedding of our model help improve its performance. When our model uses the cosine similarity of learned representations from the Siamese networks, it obtains the lowest performance in predicting food pairing scores. The concatenation (\textit{deep layer}) of two representations dramatically improves the performance of our model. This indicates that semantic relations need to be learned for predicting food pairing scores. Furthermore, the \textit{wide\&deep} architecture that learns the relation of two ingredients further boosts our model's performance. Also, utilizing the ingredient embedding for input vectors, instead of randomly initialized vectors, improves the model's performance.

\begin{table}
    \centering
    \scalebox{0.8}{
    \begin{tabular}{c|rrrrr}
    \toprule[1pt]
     & RMSE & MSE & MAE & CORR & R2 \\\midrule[1pt]
     \textbf{KitcheNette} & \textbf{0.0421} & \textbf{0.0018} & \textbf{0.0320} & \textbf{0.9249} & \textbf{0.8551}\\
     +Cosine,-Wide\&Deep & 0.0726 & 0.0054 & 0.0540 & 0.8223 & 0.5679 \\
     -Wide Layer & 0.0432 & 0.0019 & 0.0326 & 0.9206 & 0.8474 \\
     -Ingredient Embedding & 0.0501 & 0.0025 & 0.0383 & 0.8923 & 0.7958 \\\bottomrule[1pt]
    \end{tabular}
    }
    \caption{Ablation tests on the validation set.}\label{table:ablation}
\end{table}
\section{Qualitative Analysis}
For qualitative analysis, we considered to perform experiments with actual food and get human feedback, but realized it was not easy to evaluate large-scale pairing scores and beyond scope of our work. Instead, we performed various case studies. On top of that, we provide a demo version\footnote{http://kitchenette.korea.ac.kr/} of KitcheNette where anyone could retrieve the scores of ingredient pairings that they want to find.

\subsection{Finding Unknown Pairings}
To demonstrate the accuracy of KitcheNette's predictions on infrequently used food ingredient pairings, we performed a comparative analysis of the prediction results of both \textit{known} and \textit{unknown} ingredient pairings. As illustrated in Table \ref{table:unknown}, we chose three similar carbonated white wines (champagne, sparkling\_wine, and prosecco). We then calculated the score of each wine paired with a different ingredient and analyzed all the possible pairings for the three different cases below.

\textbf{Case 1} We used the champagne\&orange\_twist as a given pairing for comparison since it is a well-known ingredient pairing with a high annotated score. The orange\_twist\&sparkling\_wine and orange\_twist\&prosecco pairings do not have annotated scores since they are uncommon pairings. Additionally, we chose two other ingredients (orange\_wedge and lime\_twist) that are similar to orange\_twist, but are not frequently used with any of the three wines. As a result, we have one \textit{known} pairing and eight \textit{unknown} pairings. The prediction results of all the nine pairings were consistently high (0.33-0.45).

\textbf{Case 2} Based on their prediction scores, we paired the wines with different ingredients to create the following three unique \textit{known} pairings: champagne\&elderflower\_liqueur, sparkling\_wine\&cream\_de\_cassis, and prosecco\&lemon\_sorbet). The prediction results of the remaining six \textit{unknown} pairings were also consistently high (0.29-0.42) compared to the three given \textit{known} pairings.

\textbf{Case 3} Finally, we chose onion as it made the worst \textit{known} pairing with champagne and paired it with the remaining two wines. The predictions results of the two \textit{unknown} pairings (sparkling\_wine\&onion and prosecco\&onion) were consistently as low as the scores of the \textit{known} pairing (champagne\&onion).

In sum, our prediction results show that KitcheNette is capable of making predictions on certain pairings based on analogical reasoning which states that if A is similar to B and A forms a good pairing with C, then it is more likely that B also forms a good pairing with C. We believe that using this reasoning enhanced the performance of KitcheNette on \textit{unknown} food pairings without annotated scores.

\begin{table}
    \centering
    \small
    \scalebox{0.85}{
    \begin{tabular}{c|l|lll}
    \toprule[1pt]
     &  & champagne & sparkling\_wine & prosecco \\\midrule[1pt]
    \multirow{3}{*}{\textbf{Case 1}} & orange\_twist & 0.33$^\dagger$ & 0.39$^*$ & 0.42$^*$ \\
     & orange\_wedge & 0.37$^*$ & 0.43$^*$ & 0.45$^*$ \\
     & lime\_twist & 0.34$^*$ & 0.38$^*$ & 0.40$^*$ \\\midrule
    \multirow{3}{*}{\textbf{Case 2}} & elderflower\_liqueur & 0.34$^\dagger$ & 0.39$^*$ & 0.41$^*$ \\
     & creme\_de\_cassis & 0.29$^*$ & 0.33$^\dagger$ & 0.34$^*$ \\
     & lemon\_sorbet & 0.32$^*$ & 0.39$^*$ & 0.42$^\dagger$ \\\midrule
     \textbf{Case 3} & onion & -0.20$^\dagger$ & -0.14$^*$ & -0.17$^*$\\\bottomrule[1pt]
    \end{tabular}}
    \caption{Examples of \textit{known} and \textit{unknown} pairings. $^\dagger$ and $^*$ refer to the predicted scores of \textit{known} and \textit{unknown} pairings, respectively}\label{table:unknown}
\end{table}

\begin{table*}[!t]
    \centering
    \small
    \scalebox{0.85}{
    \begin{tabular}{c|ll|ll|ll|ll}
    \toprule[1pt]
    & \multicolumn{2}{c|}{\textbf{\large{Tomato}}} & \multicolumn{2}{c|}{\textbf{\large{Onion}}} & \multicolumn{2}{c|}{\textbf{\large{Pepper}}} & \multicolumn{2}{c}{\textbf{\large{Cinnamon}}} \\\midrule[1pt]
    \textit{Rank} & \textit{KitcheNette} & \textit{FlavorDB} & \textit{KitcheNette} & \textit{FlavorDB} & \textit{KitcheNette} & \textit{FlavorDB} & \textit{KitcheNette} & \textit{FlavorDB} \\\midrule
    1 & lettuce & tea & bay\_leaf & cocoa & oregano & ginger & allspice & pepper \\
    2 & avocado & potato & celery & garlic & ground\_beef & laurel & clove & ginger \\
    3 & cucumber & mango & ground\_beef & peanut & potato & rosemary & raisin & laurel \\
    4 & bean\_dip & guava & potato & potato & thyme & basil & baking\_soda & basil \\
    5 & eggplant & apple & carrot & tomato & elbow\_macaroni & orange & apple & rosemary \\
    6 & turmeric\_powder & grape & tomato\_paste & chive & basil & spearmint & nutmeg & nutmeg \\
    7 & garam\_masala & soybean & beef\_broth & soybean & celery & oregano & applesauce & oregano \\
    8 & red\_chili\_powder & strawberry & beef\_stock & green\_beans & onion & nutmeg & brown\_sugar & cassia \\
    9 & tostados & cocoa & green\_pepper & tea & hamburger & celery & pumpkin\_puree & tea \\
    10 & taco\_shells & mushroom & stewing\_beef & leek & marjoram & dill & canned\_pumpkin & celery\\\bottomrule[1pt]
    \end{tabular}}
    \caption{The ranking results of KitchenNette and FlavorDB\protect\cite{garg2017flavordb}.}\label{table:ranking}
\end{table*}

\begin{table*}[!t]
    \centering
    \small
    \scalebox{0.75}{
    \begin{tabular}{c|ll|ll|ll|ll}
    \toprule[1pt]
    & \multicolumn{2}{c|}{\textbf{\large{Red Wine}}} & \multicolumn{2}{c|}{\textbf{\large{White Wine}}} & \multicolumn{2}{c|}{\textbf{\large{Gin}}} & \multicolumn{2}{c}{\textbf{\large{Sake}}} \\\midrule[1pt]
    Rank & \textit{KitcheNette} & \textit{Recommendations} & \textit{KitcheNette} & \textit{Recommendations } & \textit{KitcheNette} & \textit{Recommendations} & \textit{KitcheNette} & \textit{Recommendations} \\\midrule
    1 & beef\_stock & beef & mussel & butter & angostura\_bitters & apple brandy & mirin & Japanese cusine \\
    2 & beef\_cheeks & cheeze & shad* & chicken & sweet\_vermouth & apricot brandy & katakuriko & cucumber \\
    3 & lamb\_shank & game & cockle & crab & benedictine & basil & dashi\_stock & fish \\
    4 & beef\_broth & lamb & shrimp\_shells & cream & orange\_bitters & blackberries & konnyaku & gin \\
    5 & pan\_juices* & meat, red & fish\_fumet & fish & elderflower\_liqueur & celery & burdock\_root & lemon juice \\
    6 & chicken\_backs & peper, black & lobster\_base & lobster & orange\_twist & champange & miso & salads \\
    7 & saltpeter* & steak & arborio\_rice & salmon & lemon\_twist & cilantro & soy\_sauce & sashimi and sushi \\
    8 & tomato\_paste & starawberries & skate* & scallops & simple\_syrup & cola & gochujang & shellfish \\
    9 & oxtail &  & oyster\_liquor* & shellfish & dry\_vermouth & cranberry juice & mitsuba & sugar \\
    10 & dry\_red\_wine &  & cuttlefish* & veal & orange\_wedge* & ginger & dashi & vodka \\
    11 & veal\_stock &  & shrimp &  & aquavit* & herbs & pork\_belly &  \\
    12 & ajinomoto* &  & fish\_stock &  & pisco* & lemon juice & umeboshi &  \\
    13 & pike* &  & mirlitons* &  & rye\_whiskey* & lime juice & kombu &  \\
    14 & lamb\_stock &  & crayfish* &  & honey\_syrup & mint & okonomiyaki\_sauce* &  \\
    15 & beef\_bones* &  & escargot &  & mezcal* & orange juice & bonito\_flakes &  \\
    16 & verjuice* &  & scampi* &  & absinthe & oysters & yuzu &  \\
    17 & cherry\_cola* &  & clam\_juice &  & curacao & tonic & white\_sesame\_seeds &  \\
    18 & beef\_stew\_seasoning* &  & fish\_bones &  & campari &  & wood\_ear\_mushrooms &  \\
    19 & brisket* &  & lobster\_shells* &  & lemon\_twists &  & daikon\_radish &  \\
    20 & ti\_leaves* &  & scallop &  & wheat\_starch* &  & kamaboko & \\\bottomrule[1pt]
    \end{tabular}}
    \caption{Food\&drink Pairings. The ranked pairings of KitchenNette and food\&drink recommendations from \textit{\dblquote{The Flavor Bible}}\protect\cite{page2008flavor} and \textit{\dblquote{WHAT to DRINK with WHAT you EAT}}\protect\cite{dornenburg2009drink}. The recommendations are listed in alphabetical order. $^*$ refers to the predicted scores of \textit{unknown} pairings.}\label{table:drinkfood}
\end{table*}

\subsection{Comparison of Food Pairing Ranking Results}
We performed a comparative analysis between the ingredients ranked by KitcheNette and the ranked ingredients in FlavorDB\footnote{https://cosylab.iiitd.edu.in/flavordb/search}. We selected four widely used food ingredients (tomato, onion, pepper and cinnamon) to . Then we retrieved the top 10 ingredient pairings that consist of the selected ingredients. Based on our observations, KitcheNette generally recommended food ingredients that are frequently used in everyday cooking and dining (e.g., tomato\&lettuce, onion\&ground\_beef, pepper\&oregano, cinnamon\&clove,apple). On the other hand, while FlavorDB recommended food ingredients that share a large number of chemical compounds with the selected ingredients, some of the recommendations did not pair well with the selected ingredients (e.g., tomato\&tea, onion\&cocoa, pepper\&orange) for cooking and dining.
  
\subsection{Discovering New Drink-Food Pairings}
We also found that KitcheNette can discover new food-drink pairings, which we believe is one of the main aims of food pairing. As illustrated in Table~\ref{table:drinkfood}, we compared our model's food\&drink recommendations with those from \textit{\dblquote{The Flavor Bible}}~\protect\cite{page2008flavor} and \textit{\dblquote{WHAT to DRINK with WHAT you EAT}}~\protect\cite{dornenburg2009drink}. We found that KitcheNette not only provides recommendations that are consistent with the recommendations of culinary experts from the books. books but also recommends far more pairings than the two books. 

For red wine and white wine, our model recommended a variety of meat (e.g., beef, lamb) and specific seafood ingredients (e.g., mussel, lobster, shrimp) respectively. Our model recommended authentic Japanese food ingredients to pair with sake, which shows that our data-driven learning model is also able to recommend food ingredients less common in non-Asian cuisines.
\section{Conclusion \& Future Work}
In this work, we introduced KitchenNette which predicts food ingredient pairing scores based on a large amount of food recipe data, and ranks food ingredient pairings based on the predicted scores. Our model which has Siamese deep neural networks is trained on dataset containing more than 300k food ingredient pairing scores. We demonstrate that our model discovers new and unknown pairings and achieves better ranking results than the existing food pairing ranking models. Also, our model discovers new drink-food pairings and accurately predicts the scores of new food ingredient pairings.

For future work, we plan to use a graph-based neural network architecture to train on one-to-many ingredient pairings, instead of on one-to-one pairings, which were used by our model's Siamese networks. Also, we plan to add the chemical information of food ingredients to the ingredient embeddings and use more detailed information on food ingredients from food encyclopedias. Last, we would like to use more novel and authentic recipes to help our model to recommend more versatile food ingredient pairings.


\clearpage
\newpage

\bibliographystyle{named}
\bibliography{ijcai19}

\appendix

\end{document}